\documentclass[runningheads]{llncs}
\usepackage[T1]{fontenc}

\usepackage{multirow}
\usepackage{booktabs}
\usepackage{amsmath}
\usepackage[inkscapeformat=png]{svg}
\usepackage[misc]{ifsym}
\usepackage[colorlinks=true,
            linkcolor=blue,     
            citecolor=blue,     
            urlcolor=blue       
           ]{hyperref}
\begin{document}
\title{Embedding-Based Federated Data Sharing via Differentially Private Conditional VAEs}

\titlerunning{Embedding-Based Federated Data Sharing via DP-CVAE}

\author{Francesco Di Salvo\textsuperscript{* (\Letter)} \and Hanh Huyen My Nguyen\textsuperscript{*} \and Christian Ledig}
\authorrunning{F. Di Salvo et al.}
%
\institute{xAILab Bamberg, University of Bamberg, Germany \\
\email{francesco.di-salvo@uni-bamberg.de}\\
}

\maketitle              
%

\renewcommand{\thefootnote}{}
\footnotetext{
\textsuperscript{*} Equal contribution. Code available at 
\href{https://github.com/myng15/federated-dp-cvae}{github.com/myng15/federated-dp-cvae}.}

\begin{abstract} 

Deep Learning (DL) has revolutionized medical imaging, yet its adoption is constrained by data scarcity and privacy regulations, limiting access to diverse datasets. Federated Learning (FL) enables decentralized training but suffers from high communication costs and is often restricted to a single downstream task, reducing flexibility. We propose a data-sharing method via Differentially Private (DP) generative models. By adopting foundation models, we extract compact, informative embeddings, reducing redundancy and lowering computational overhead. Clients collaboratively train a Differentially Private Conditional Variational Autoencoder (DP-CVAE) to model a global, privacy-aware data distribution, supporting diverse downstream tasks. Our approach, validated across multiple feature extractors, enhances privacy, scalability, and efficiency, outperforming traditional FL classifiers while ensuring differential privacy. Additionally, DP-CVAE produces higher-fidelity embeddings than DP-CGAN while requiring $5{\times}$ fewer parameters. 


\end{abstract}
\section{Introduction}

Deep Neural Networks (DNNs) have driven remarkable advancements in medical imaging, yet their adoption in clinical practice remains constrained by limited data availability and stringent privacy requirements \cite{stacke2020measuring}. Medical datasets are siloed across institutions, and low-prevalence diseases further limit the availability of diverse, high-quality training data \cite{litjens2017survey}. While collaborative data sharing could mitigate these challenges \cite{shilo2020axes}, strict privacy regulations (\textit{e.g.}, HIPAA, GDPR) make centralized dataset aggregation infeasible. To address these constraints, Federated Learning (FL) \cite{mcmahan2017communication} has emerged as a privacy-preserving alternative, allowing institutions to collaboratively train models without sharing raw data. A widely adopted strategy, FedAvg \cite{mcmahan2017communication}, aggregates model updates from participating clients to construct a global model. However, FL introduces several challenges. Communication overhead remains high, particularly when deploying deep architectures such as Vision Transformers (ViTs) \cite{dosovitskiy2020image}, which significantly increases transmission costs. Furthermore, FL is typically restricted to a single downstream task (\textit{e.g.,} classification, segmentation), limiting generalizability. 

To mitigate transmission costs, most FL research focuses on lightweight architectures, which often come at the expense of model performance and robustness. An alternative to model-sharing is data-sharing via privacy-preserving synthetic data generation, which reduces communication overhead while enabling broader downstream applications \cite{koetzier2024generating,ktena2024generative}. Several works have explored federated generative models for this purpose (see \cite{gargary2024systematic} for a comprehensive review). However, generating realistic, task-relevant synthetic data remains challenging, as it often requires a large number of diverse training samples to ensure fidelity to the original distribution. While Generative Adversarial Networks (GANs) \cite{goodfellow2014generative} and diffusion models \cite{ho2020denoising} achieve high-fidelity image synthesis, they exhibit notable limitations \cite{koetzier2024generating}. GANs suffer from mode collapse, producing low-diversity synthetic data, while diffusion models are computationally expensive and exhibit high latency, making them impractical for resource-constrained federated environments. In contrast, Variational Autoencoders (VAEs) and Conditional VAEs (CVAEs), despite producing lower-fidelity images, \textit{e.g.,} blurred reconstructions, offer notable advantages. In fact, they avoid mode collapse while being more computationally efficient than GANs and diffusion models. 

\begin{figure}[h]
	\centering
        \includegraphics[width=1.0\textwidth]{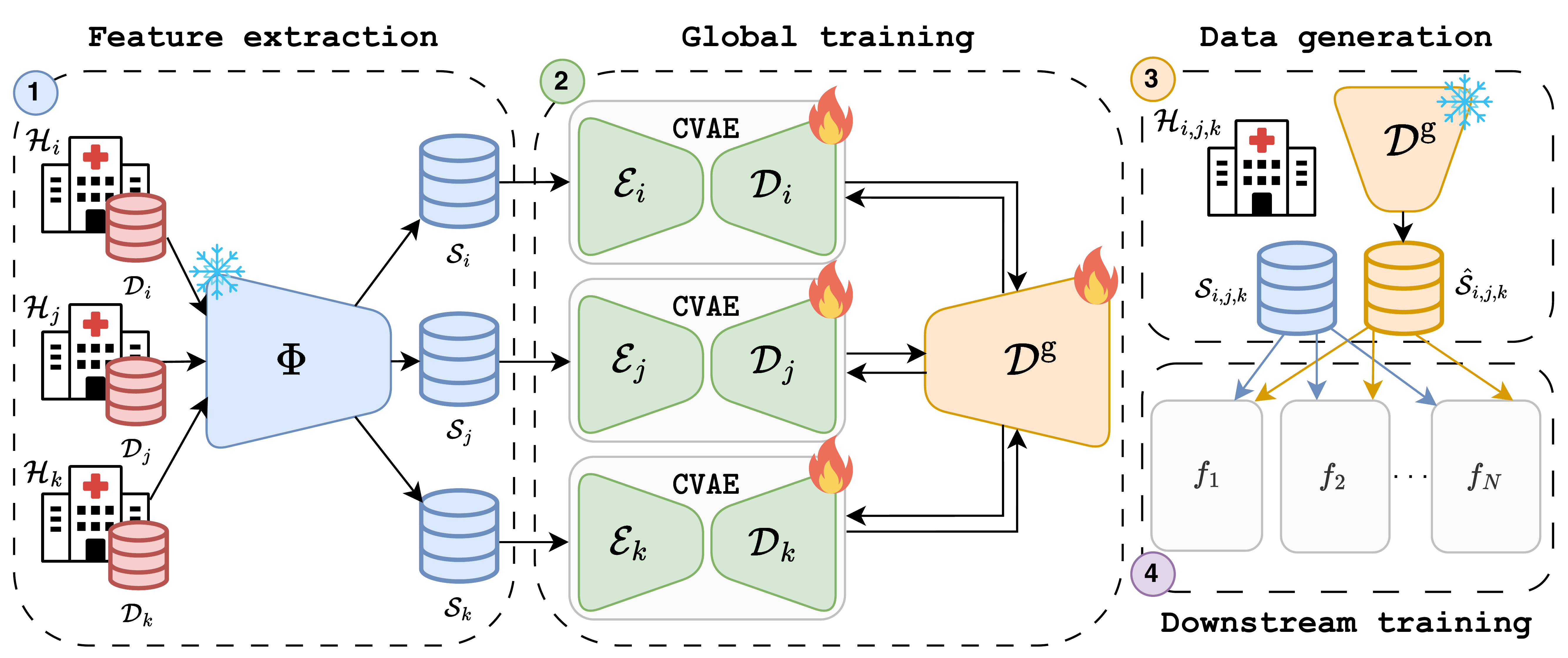}
        \caption{\label{fig:pull_figure}Illustration of our proposed methodology. (1) Each client $\mathcal{H}$ encodes its image-based dataset $\mathcal{D}$ into an embedding-based dataset $\mathcal{S}$ using a large, pre-trained foundation model $\mathrm{\Phi}$, reducing data storage requirements and computational overhead. (2) Clients collaboratively train a lightweight DP-CVAE ($\mathcal{E},\mathcal{D}$) and periodically share decoder weights, which are aggregated into a global decoder $\mathcal{D}^{\text{g}}$. This shared decoder captures cross-client variation while preserving local data privacy. (3) Each client independently generates a synthetic dataset $\hat{\mathcal{S}}$ using the globally trained generative model, and (4) utilizes (real) local and (synthetic) global data for any downstream task $f$.}
\end{figure}

While VAEs and CVAEs have been explored in federated settings, prior work has primarily applied them to simpler generative tasks, such as MNIST-like datasets \cite{pfitzner2022dpd}, sensor data \cite{10217998}, or joint training with a downstream classifier \cite{10119117}, limiting their adaptability. A recent work has demonstrated that generating synthetic feature embeddings using a CVAE preserved classification performance, comparable to real embeddings, while enhancing data privacy \cite{disalvo2024privacy}. A key enabler of this approach is the use of foundation models \cite{oquab2024dinov}, known to be robust to domain shifts \cite{paul2022vision}. Furthermore, these models produce compact and diagnostically relevant feature representations while reducing redundancy in raw images and enabling low-cost downstream learning. Training a CVAE on feature embeddings rather than raw images allows to better capture feature distributions, making it less susceptible to fidelity degradation. This motivates our extension of CVAE to a federated setting, where, as illustrated in Figure \ref{fig:pull_figure}, clients collaboratively train a differentially-private global generative model. Unlike prior FL settings, our approach decouples generative modeling from task-specific constraints, allowing greater flexibility across applications. In summary, our contributions are: 

\begin{itemize}
    \item We propose a lightweight federated generative model with differential privacy to address data scarcity and enable privacy-preserving data sharing in medical image analysis. Our approach decouples data-sharing from downstream tasks, enhancing generalizability and adaptability across applications.
    \item We empirically demonstrate that our federated generative approach and subsequent downstream training outperform traditional federated classifiers across multiple datasets, achieving higher balanced accuracy.
    \item We show that training a lightweight CVAE on feature embeddings achieves higher fidelity than GAN-based approaches while requiring approximately $5{\times}$ fewer parameters, significantly improving computational efficiency.
\end{itemize}

\section{Method}

\subsection{Feature extraction}
Traditional FL pipelines often use lightweight architectures \cite{wu2023fediic,xia2024enhancing} to mitigate communication overhead and computational demands. However, this may compromise robustness and generalization, which are essential in medical image analysis for capturing high-quality feature representations. To address these limitations, we utilize a (shared) large pre-trained foundation model to extract compact, informative feature embeddings while substantially reducing the inherent information redundancy of raw images. While any feature extractor can be used, in this work, we adopt DINOv2 Base \cite{oquab2024dinov} due to its state-of-the-art performance across different domains and tasks. 
Considering a federation of $M$ clients, each client $m$ holds a private image-based dataset $\mathcal{D}_m := \{ (\mathbf{d}_i^m, y_i^m) \}_{i=1}^{n_m}$, where $\mathbf{d}_i^m$ represents the $i$-th raw image and $y_i^m \in \mathcal{Y}$ is its corresponding label. After applying feature extraction locally using a pre-trained foundation model, the raw images are transformed into feature embeddings, resulting in the feature-based dataset $\mathcal{S}_m := \{ (\mathbf{x}_i^m, y_i^m) \}_{i=1}^{n_m}$ \noindent where $\mathbf{x}_i^m \in \mathcal{X}_m$ is the feature embedding of $\mathbf{d}_i^m$.

\subsection{Federated- and differentially-private generative model}

\subsubsection{CVAE}

A recent study \cite{disalvo2024privacy} demonstrated that generating synthetic training sets at the embedding level, rather than raw images, introduces privacy preservation while observing only minimal performance degradation in downstream classification. Following this approach, we adopt a CVAE with a symmetric architecture, consisting of three linear layers for both the class-conditional encoder and decoder. The model is trained by minimizing a reconstruction loss (Mean Squared Error) while enforcing latent space normality through a Kullback-Leibler (KL) divergence loss with a standard normal prior.

\subsubsection{Differential Privacy (DP)}

To enforce formal privacy guarantees, we integrate Differential Privacy \cite{abadi2016dpsgd,dwork2006differential}, ensuring that the synthetic embeddings remain indistinguishable with respect to the presence or absence of any single data point in the original dataset. This is achieved by bounding the sensitivity of the generative model to individual samples, thereby preventing adversaries from reconstructing or inferring specific data points from the released synthetic dataset. DP-CVAE ensures that the posterior distribution of the generated embeddings remains statistically similar irrespective of whether a given sample is included in the training data. We enforce $(\epsilon, \delta)$-differential privacy by introducing calibrated noise into the generative process. Formally, a mechanism $\mathcal{M}$ applied to the private dataset $\mathcal{S}_m$ is $(\epsilon, \delta)$-differentially private if, for any two neighboring datasets $\mathcal{S}_m$ and $\mathcal{S}_m'$ differing by at most one sample, and for any possible output $\mathcal{\hat{S}}_m$:

\begin{equation}
\Pr[\mathcal{M}(\mathcal{S}_m) \in \mathcal{\hat{S}}_m] \leq e^\epsilon \Pr[\mathcal{M}(\mathcal{S}_m') \in \mathcal{\hat{S}}_m] + \delta, \quad \epsilon > 0, \delta \in [0, 1)
\end{equation}

where $(\epsilon, \delta)$ is an upper bound on the privacy loss between before and after an individual was added to the dataset, giving us a formal privacy guarantee. To achieve this, we integrate DP-SGD \cite{abadi2016dpsgd} into the CVAE training process, where noise $\mathcal{N}(0, \sigma^2)$ is added to the \textit{per-sample} gradients, ensuring that each model update is differentially private. Before adding noise, we clip gradients so that they have a maximum $\ell_2$ norm of $C$ to limit the influence of a single data point. Notably, each client performs DP locally on their CVAE before sending the local decoder to the server to update the global CVAE decoder. Any outputs of the resulting CVAE perturbed by the DP noise are guaranteed to protect an individual's data used in training according to the chosen $\epsilon$. However, this noise comes at the cost of utility loss, which grows as $\epsilon$ decreases.

\subsubsection{Model aggregation and data generation} Each client maintains a personalized encoder, which adapts to its local data distribution, mapping samples into a shared latent space. Meanwhile, the decoders are jointly trained using a Federated Averaging (FedAvg) strategy, ensuring that the aggregated decoder learns a globally representative feature reconstruction function while remaining compact and efficient. Formally, at each communication round $t$, the server aggregates the decoder weights $\theta_{\text{dec}}^m$ from all $M$ participating clients as follows:

\begin{equation} \theta_{\text{dec}}^{(t+1)} = \sum_{m=1}^{M} w_m \theta_{\text{dec}}^{m, (t)}, \quad w_m=\frac{ n^{\text{train}}_m}{\sum_{m=1}^M n^{\text{train}}_m} \end{equation}
 
where $\theta_{\text{dec}}^{m, (t)}$ represents the decoder parameters of client $m$ at round $t$, and $\theta_{\text{dec}}^{(t+1)}$ represents the updated global decoder parameters distributed back to the clients. Once the federated training of the global decoder is completed, each client utilizes it to generate a synthetic dataset that approximates the global data distribution while ensuring privacy preservation. Given a target synthetic dataset size $N$, each client constructs its global dataset $\hat{\mathcal{S}}_m$ as follows:

\begin{equation} 
    \hat{\mathcal{S}}_m = \{ (\hat{\mathbf{x}}_i, \hat{y}_i) \}_{i=1}^{N}, \quad  \quad \hat{\mathbf{x}}_i = \text{Dec}_{\theta_{\text{dec}}^{(t+1)}}(\mathbf{z}_i \mid \hat y_i)
\end{equation}

where $\mathbf{z}_i \sim \mathcal{N}(\mathbf{0}, \mathbf{I})$ is sampled from a standard normal Gaussian distribution, $\hat{y}_i \sim C$ is sampled from a selected class distribution, and $\text{Dec}_{\theta_{\text{dec}}^{(t+1)}}$ is the globally trained decoder obtained after FedAvg aggregation. 

\subsection{Downstream classification}

In our setup, each client has access to both a local (private) dataset and a global dataset, the latter generated using the shared global decoder. This setup enables clients to train models tailored to their specific downstream tasks. For instance, clients can utilize the same synthetic data to train classifiers with different label granularities, model data distributions for anomaly detection,  address out-of-distribution (OOD) detection problems, and more. \newline

We consider image classification as the primary downstream task for a standardized evaluation. The availability of both local and global datasets enables \textit{personalized} FL, in which each client trains a personalized model while still benefiting from other clients' knowledge. For example, kNN-Per \cite{marfoq2022personalized} uses client-level local memorization to improve individual performance but still jointly learns the global representations, and FedRep \cite{collins2021exploiting} aims to learn a shared feature representation across clients. In contrast, we take advantage of the shared feature representations of a foundation model and train for each client one model on the local data and another on the global (synthetic) data for the downstream task (instead of for representation learning). The final classification prediction is obtained through a weighted interpolation between these two models, controlled by a tunable parameter $\lambda_m$, which is optimized based on the validation set to balance local specialization and global generalization. For a test sample $\mathbf{x}_{\text{test}}$, the interpolated probability distribution is computed as:

\begin{equation} 
\label{eq:weighted_prob}
    P_{\lambda_{m}}(y \mid \mathbf{x}_{\text{test}}) = \lambda_m P_{\text{local}}(y \mid \mathbf{x}_{\text{test}}) + (1 - \lambda_m) P_{\text{global}}(y \mid \mathbf{x}_{\text{test}})
\end{equation}

where $\lambda_{m} \in [0,1]$ determines the trade-off between personalization and global knowledge. The final predicted class $\hat{y}_\text{test}$ is given by:

\begin{equation}\hat{y}_{\text{test}} = \arg\max_{c \in \mathcal{Y}} P_{\lambda_{m}}(y \mid \mathbf{x}_{\text{test}})
\end{equation}

Intuitively, as $\lambda_{m} \to 1$, the model prioritizes local data, utilizing client-specific knowledge. Conversely, as $\lambda_{m} \to 0$, the model relies more on globally shared information, benefiting from knowledge aggregated across the federation.

\section{Experimental results}

\subsection{Experimental settings}

\subsubsection{Datasets and metrics} 

We evaluate the downstream classification performance of our method across two distinct classification settings. Following \cite{li2024sift}, we use the Abdominal CT dataset (Sagittal view) \cite{xu2019efficient}, utilizing the splits from \cite{medmnistv2}. It presents $25,211$ images across $11$ classes. We distribute the data among $10$ clients, simulating IID conditions, together with highly non-IID conditions using a Dirichlet distribution ($\alpha = 0.3$). Furthermore, as in \cite{chen2023fedsoup}, we use a subset of $4{,}600$ images from Camelyon17-Wilds \cite{koh2021wilds}, a binary dataset consisting of histopathological images from five hospitals, treated as five individual clients. Each local dataset is further split into train--val--test ($60{:}20{:}20$). While Camelyon17-Wilds is class-balanced, the CT dataset is imbalanced, therefore, we report the overall mean and standard deviation across clients of accuracy and balanced accuracy. 

\subsubsection{Implementation details} 
Both the federated and locally trained classifiers are implemented as single-layer linear models, a standard approach for evaluating foundation model embeddings \cite{oquab2024dinov}. Both the FedAvg classifier and our DP-CVAE are trained for $50$ communication rounds, with $5$ local epochs per round, using the SGD optimizer with a learning rate of $1{\times}10^{-3}$. We apply DP to our CVAE and, for comparison, to a Conditional GAN (CGAN), using the OPACUS library \cite{opacus}, with $(\epsilon, \delta) = (1.0, 1\times10^{-4})$ and a clipping norm of 1.5. This choice, with $\epsilon \leq 1$ and $\delta \ll 1/n$, where $n$ is the average number of training samples per client in our experiments, ensures meaningful privacy guarantees while maintaining data utility \cite{lange2022privacy,nasr2021adversary}. The interpolation parameter $\lambda_m$ (\textit{c.f.} Equation \ref{eq:weighted_prob}) is automatically selected from $\{0.0, 0.1, \dots, 1.0\}$ based on validation set performance.  Lastly, our downstream classifiers are trained separately on local and global data, using Adam optimizer with a learning rate of $1{\times}10^{-3}$ for $100$ epochs. For comparison, we evaluate FedAvg and its enhanced version, FedProx \cite{li2020federated}, which addresses non-IID data by adding a regularization term that penalizes large deviations from the global model. Additionally, as a direct competitor to our approach, we include ``FedLambda'', our adapted version of kNN-Per \cite{marfoq2022personalized}, where the local kNN component is replaced with a local linear classifier and the global classifier is trained through FedAvg.

\begin{table}
\caption{Accuracy (ACC) and balanced accuracy (BACC), averaged across clients over three seed runs, between federated classifiers and our proposed data-sharing and classification method. We highlight in \textbf{bold} the top two methods.}\label{tab:tab1} 
\centering
\begin{tabular}{lccccccccccccccccc}
\toprule


     & \multicolumn{2}{c}{CT (IID)}
     & \multicolumn{2}{c}{CT ($\alpha=0.3$)}
     & Camelyon \\ 
     \cmidrule(l){2-3} \cmidrule(l){4-5} \cmidrule(l){6-6} 

     & ACC & BACC
     & ACC & BACC
     & ACC  \\ \midrule

FedAvg
     & 73.33\scriptsize{$\pm$1.14}
     & 67.00\scriptsize{$\pm$1.15}
     & 64.85\scriptsize{$\pm$5.88}
     & \textbf{58.74\scriptsize{$\pm$2.93}}
     & 90.62\scriptsize{$\pm$2.34}
     \\ 
     
FedProx
     & 73.26\scriptsize{$\pm$1.19}
     & 66.95\scriptsize{$\pm$1.17}
     & 64.76\scriptsize{$\pm$6.13}
     & 58.66\scriptsize{$\pm$5.84}
     & 90.65\scriptsize{$\pm$2.01}
     \\ 

FedLambda
     & 77.21\scriptsize{$\pm$0.80}
     & 71.32\scriptsize{$\pm$0.91}
     & 81.03\scriptsize{$\pm$3.83}
     & \textbf{59.08\scriptsize{$\pm$2.54}}
     & 92.54\scriptsize{$\pm$1.03}
     \\ \midrule
     
DP-CGAN
     & \textbf{77.60\scriptsize{$\pm$1.36}}
     & \textbf{71.94\scriptsize{$\pm$1.18}}
     & \textbf{88.84\scriptsize{$\pm$1.99}}
     & 57.49\scriptsize{$\pm$2.41}
     & \textbf{93.95\scriptsize{$\pm$0.98}}
     \\

DP-CVAE
     & \textbf{77.54\scriptsize{$\pm$0.76}}
     & \textbf{71.84\scriptsize{$\pm$0.82}}
     & \textbf{88.94\scriptsize{$\pm$1.35}}
     & 57.58\scriptsize{$\pm$3.33}
     & \textbf{94.49\scriptsize{$\pm$1.28}} 
     \\ 

\bottomrule
\end{tabular}
\end{table}

\subsection{Downstream classification}

Table \ref{tab:tab1} presents the classification results across different experimental settings. Overall, federated data-sharing schemes using generative models outperform federated classifiers, with DP-CVAE achieving the highest performance in most cases. Notably, incorporating the local predictions under personalized FL settings leads to a substantial improvement over standard FedAvg, highlighting the benefits of a personalized adaptation. Furthermore, it is important to note that these federated classifiers (including FedLambda) are not differentially private, whereas our approach enforces differential privacy guarantees, which may have a slight impact on performance. In the heterogeneous CT setting ($\alpha = 0.3$), our method achieves comparable or slightly lower balanced accuracy than federated classifiers, yet it substantially outperforms them in terms of accuracy. The most notable performance gap between our generative-based approach and federated classifiers is observed on Camelyon17, where each client has an average of solely $500$ train samples. This result underscores that our generative method remains highly effective even under limited-data conditions. Furthermore, while DP-CGAN and DP-CVAE achieve comparable performance overall, DP-CVAE notably outperforms DP-CGAN on Camelyon17, suggesting that CVAE is less data-hungry, making it particularly suited for limited-data availability scenarios.

\subsection{Ablation studies}

\subsubsection{Choice of the generative model} This ablation study compares DP-CVAE and DP-CGAN in their computational efficiency and ability to preserve the original data distribution. We compute the average Wasserstein distance ($\mathcal{W}$) between each client's real dataset and the synthetic dataset they generate using the globally trained DP-CGAN or DP-CVAE decoder. As shown in Figure \ref{fig:wass}, DP-CVAE consistently achieves a lower $\mathcal{W}$, indicating better fidelity. Furthermore, DP-CVAE requires $5{\times}$ fewer parameters than DP-CGAN, thereby enhancing efficiency and reducing latency, making it more suitable for federated learning. 

\subsubsection{Generalizability and robustness} 
\begin{figure}[htb]
  \centering
  \begin{minipage}[b]{0.4\textwidth}
    \centering
    \includegraphics[width=0.8\linewidth]{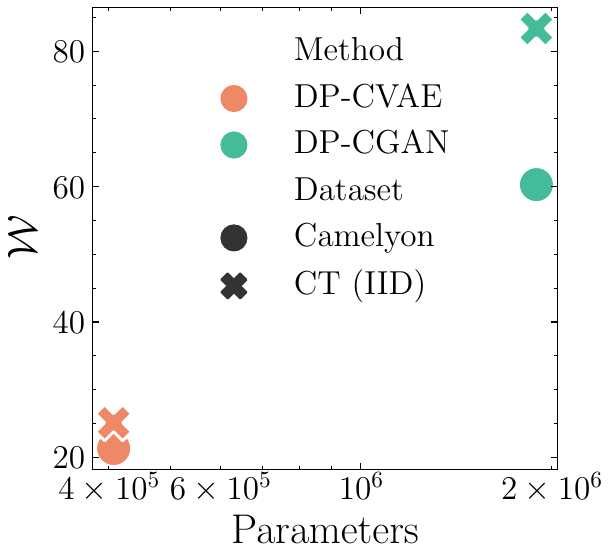}
    \caption{Reconstruction fidelity, measured by the Wasserstein distance ($\mathcal{W} \downarrow$) between real and synthetic samples, comparing DP-CVAE and DP-CGAN. Results are analyzed in relation to the number of model parameters.}
    \label{fig:wass}
  \end{minipage}
  \hfill
  \begin{minipage}[b]{0.55\textwidth}
    \centering
    \includegraphics[width=1.0\linewidth]{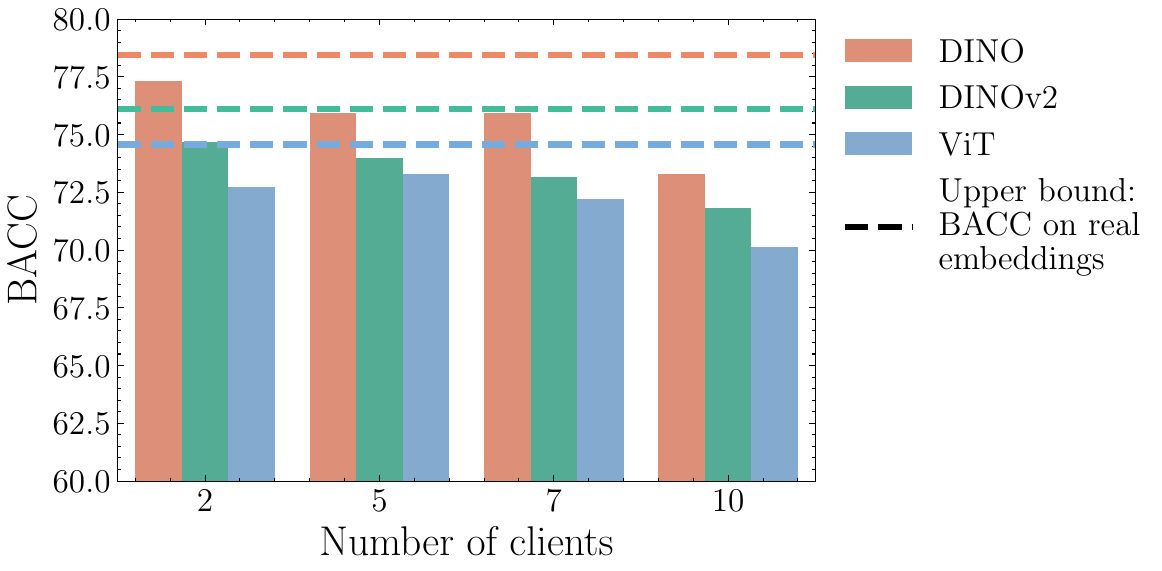}
    \caption{Generalizability of our method across different backbones and increasing numbers of clients (\textit{i.e.}, fewer samples per client), utilizing CT (IID). The dashed lines represent the BACC obtained when training a linear classifier on real image embeddings with the original train--val--test split.}
    \label{fig:utility}
  \end{minipage}
\end{figure}

Utilizing CT (IID) as a reference, we demonstrate in Figure \ref{fig:utility} that our approach generalizes effectively across different (Small) backbones, including DINOv2, DINO, and ViT, maintaining consistent classification performance. Additionally, as the number of clients increases, leading to fewer samples per client, it exhibits a minimal performance drop. This suggests that our approach is data-efficient and remains robust across varied feature representations and dataset sizes.

\section{Discussion}

\subsubsection{Limitations and future work} 

Although generative models trained on embedding spaces are more efficient and less data-demanding, they remain susceptible to data and label imbalance, a well-known challenge in federated learning. This limitation can be addressed by incorporating techniques from long-tailed learning \cite{10105457}, enhancing robustness across imbalanced distributions. Furthermore, our current generative model samples embeddings with a fixed unit variance. Introducing learned or class-specific variance parameters could improve the quality and expressiveness of synthetic embeddings, leading to better downstream performance. Finally, conditioning the generative model on additional confounders (\textit{e.g.}, domain-specific attributes) could further enhance data diversity and mitigate inherent biases in the training distribution, improving generalization and fairness in real-world applications.

\subsubsection{Conclusions} 

This work introduces a federated learning method that shifts from traditional (downstream) model-sharing to privacy-preserving data-sharing, utilizing DP-CVAEs trained on foundation model embedding spaces. Unlike conventional FL approaches that are constrained to a single downstream task, our method enables flexible and adaptive synthetic data generation, allowing clients to tailor their datasets for diverse applications. Through comprehensive experiments on medical imaging datasets, we demonstrated that our approach outperforms federated classifiers, achieving substantially higher accuracy. Additionally, we showed that training a lightweight CVAE on feature embeddings preserves data fidelity more effectively than GANs, while requiring significantly fewer parameters. These findings position our approach as a promising alternative for enabling secure, flexible, and high-performance FL in medical image analysis.


\begin{credits}
\subsubsection{\ackname} This study was funded through the Hightech Agenda Bayern (HTA) of the Free State of Bavaria, Germany.

\subsubsection{\discintname} The authors have no competing interests to declare that are relevant to the content of this article. 
\end{credits}

%
%
%
\bibliographystyle{splncs04}
\bibliography{bibliography}
\end{document}